\documentclass[10pt,twocolumn,letterpaper]{article}

\usepackage{pgfplots}
\usepackage{wacv}
\usepackage{times}
\usepackage{epsfig}
\usepackage{graphicx}
\usepackage{amsmath}
\usepackage{amssymb}

\usepgfplotslibrary{colorbrewer}
\usepackage{subfig}
\usepackage{array}
\usepackage{booktabs}
\usepackage{multirow}
\usepackage{balance}
\pgfplotsset{cycle list/Set1-7}
\newcolumntype{R}[1]{>{\raggedleft\let\newline\\\arraybackslash\hspace{0pt}}m{#1}}

\wacvfinalcopy 

\def\wacvPaperID{395} 

\ifwacvfinal\fi
\begin{document}

\title{Shape-only Features for Plant Leaf Identification}

\author{Charlie Hewitt \\
University of Cambridge\\
{\tt\small ac@chewitt.me}
\and
Marwa Mahmoud \\
University of Cambridge\\
{\tt\small marwa.mahmoud@cl.cam.ac.uk}
}

\maketitle
\ifwacvfinal\thispagestyle{empty}\fi

\begin{abstract}
This paper presents a novel feature set for shape-only leaf identification motivated by real-world, mobile deployment. The feature set includes basic shape features, as well as signal features extracted from local area integral invariants (LAIIs), similar to curvature maps, at multiple scales. The proposed methodology is evaluated on a number of publicly available leaf datasets with comparable results to existing methods which make use of colour and texture features in addition to shape. Over 90\% classification accuracy is achieved on most datasets, with top-four accuracy for these datasets reaching over 98\%. Rotation and scale invariance of the proposed features are demonstrated, along with an evaluation of the generalisability of the approach for generic shape matching.
\end{abstract}

\section{Introduction}
Shape matching and classification is a long-standing problem in computer vision and has a huge variety of uses, from tracking hands in videos~\cite{chamfer} to efficient OCR~\cite{shape-ocr}. With the increasing deployment of computer vision systems in end-user applications, the time and space efficiency as well as the adaptability of models are becoming an increasing concern. As such, compact yet descriptive representations of shapes for use in shape matching and classification applications are particularly useful.

In this paper we tackle the problem of leaf identification, a particularly challenging subset of shape matching. Intra-class variation of leaf shapes is typically quite large, while inter-class variation may be very small. Automatic leaf identification greatly speeds up the process of collection, cataloguing and monitoring for botanical researchers, a task which can take significant time for experts to perform and may be impossible for amateurs.

For this type of image classification task, one might be tempted to turn to deep learning approaches, but there are two key issues with this approach. Firstly, there is no control over the features which the CNN learns and, as such, non-relevant factors such as lighting conditions and background objects (e.g., scale rulers in lab images) might be considered when they are clearly not relevant to the actual problem of leaf identification. Secondly, large CNNs required vast datasets and can take many days to train, as well as having large storage requirements for deployment. This makes these approaches unfeasible for any practical application where storage may be limited and where data for leaf classes may be sparse and new data might need to be incorporated for deployment.

Many existing leaf identification techniques rely on a fusion of feature types, including shape, colour and texture. Colour features often prove quite effective in increasing accuracy on leaf datasets which typically include leaves collected only during summer. For real-world leaf classification, where colours of leaves from the same tree will vary drastically across the year and under different lighting conditions, these colour features are rendered almost completely obsolete. Texture and vein features rely on very high quality images, often aided by back-lighting. Obtaining these features from lower quality field images is unrealistic.

We, therefore, propose the use of a novel shape-only feature set for leaf identification, requiring only an object segmentation as input. This feature set combines a number of statistical and spectral features extracted from LAII at multiple resolutions, in addition to conventional, holistic shape features. Similar to curvature maps, LAIIs have a number of desirable attributes for shape representation; they are translation, rotation and scale invariant, as we demonstrate experimentally. Evaluation of the proposed shape-only features on popular leaf identification datasets demonstrates comparable results to previous work which utilises more complex feature sets. The proposed features have also proved to be generalisable to other generic shape matching domains. Since classifiers used in the proposed methodology are typically lightweight, this makes our methodology quickly updatable and easily deployable in real-world scenarios.


\section{Related Work}
Shape matching is one of the most common problems in computer vision and many methods have been proposed to tackle this task with varying success. Most methods focus on extracting some form of shape descriptors which capture the key elements of the shape irrespective of its transformation within the scene, such as shape descriptors~\cite{shapematching}, chamfers~\cite{chamfer} and complex networks~\cite{shapecn}. These methods are often good at capturing large shape features and therefore useful in identifying common objects which typically vary significantly in shape, but are less effective at capturing small variations such as serrated leaf margins.

An alternative method is to represent the shape completely as a 1D curvature map which is then used for matching; this method has been suggested for generic shape matching~\cite{curvature}, though has not been particularly popular. This is perhaps due to the higher computational demand compared with other methods which typically achieve similar results for generic shapes. An improvement over curvature maps are LAII~\cite{integral}. These are largely analogous to curvature maps, though with the additional property of retaining locality of features over scale, and are the representation used in this paper.

Leaf identification is also a well-established problem in computer vision, again with a vast array of proposed techniques. Most methods use basic shape features including solidity, circularity, convexity and eccentricity~\cite{leafprocessor,costarica,foliage,flavia,glcmshen,shapecoltex,neural}. Some also use more complex shape features such as Zernike moments~\cite{foliage}, bending energy~\cite{leafprocessor}, histograms of curvature scale (also derived from LAIIs)~\cite{leafsnap,costarica}, centroid radial maps~\cite{folio} and Shen features~\cite{glcmshen}. Mallah \emph{et al.}~\cite{100leaves} also include features specifically targeted at the leaf margin to represent this information specific to leaf classification.

Colour features have been used, along with texture and vein features typically based on grey level co-occurrence matrix features (GLCM)~\cite{foliage,shapecoltex,glcmshen,neural,folio}. Local binary variance patterns (LBVP) can also provide a basis for texture features~\cite{costarica}.

Deep learning approaches have been attempted with the end-to-end training of large CNNs~\cite{leafnet,bagofwords,treelogy}, though most methods follow a classical machine learning pipeline, most often using SVM or kNN classifiers, with some investigation into neural network classifiers~\cite{flavia,neural,foliage}.

\section{Leaf Datasets}
\begin{figure}
    \captionsetup[subfloat]{justification=centering}
    \centering
    \null{}\hfill
    \subfloat[Flavia]{%
        \centering
        \includegraphics[height=0.2\linewidth]{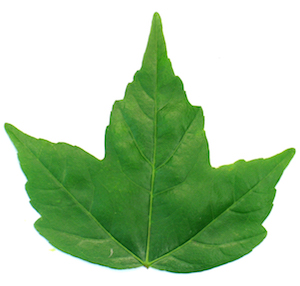}
    }
    \hfill
    \subfloat[Foliage]{%
        \centering
        \includegraphics[height=0.2\linewidth]{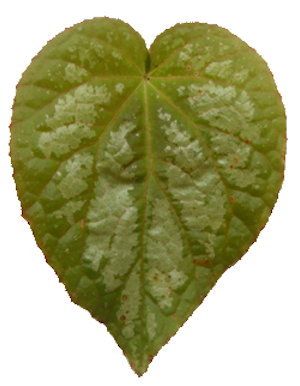}
    }
    \hfill
    \subfloat[Folio]{%
        \centering
        \includegraphics[height=0.2\linewidth]{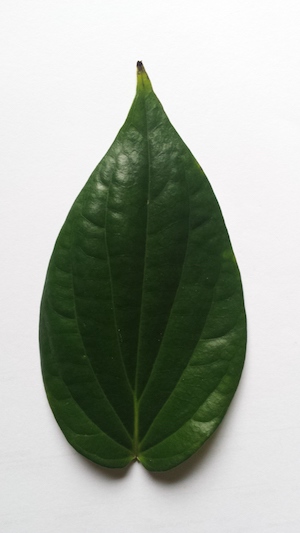}
    }
    \hfill
    \subfloat[Leafsnap Field]{%
        \centering
        \includegraphics[height=0.2\linewidth]{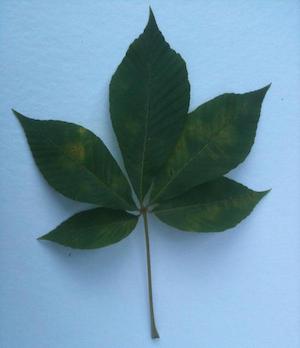}
    }
    \hfill\null{}
    \\
    \null{}\hfill
    \subfloat[Leafsnap Lab]{%
        \centering
        \includegraphics[height=0.2\linewidth]{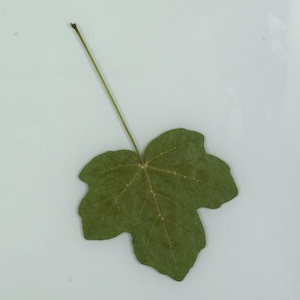}
    }
    \hfill
    \subfloat[ShapeCN]{%
        \centering
        \includegraphics[height=0.2\linewidth]{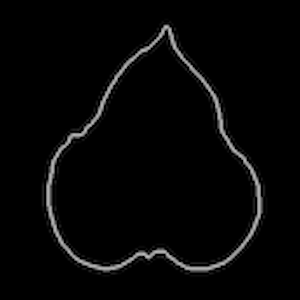}
    }
    \hfill
    \subfloat[Swedish Leaves]{%
        \centering
        \includegraphics[height=0.2\linewidth]{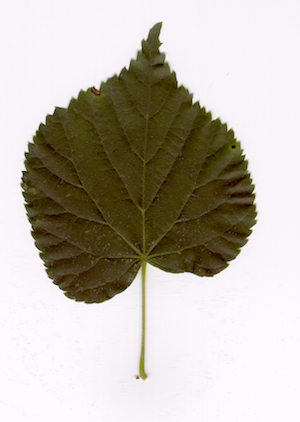}
    }
    \hfill
    \subfloat[100-Leaves]{%
        \centering
        \includegraphics[height=0.2\linewidth]{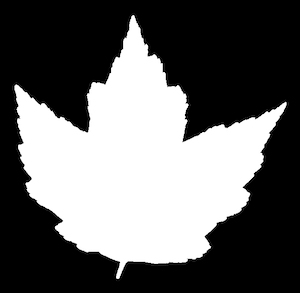}
    }
    \hfill\null{}
    \caption{Example images from standard leaf datasets demonstrating the large variability between datasets.\label{fig:datasets}}
\end{figure}

A number of publicly available leaf datasets are used to provide a good indication of the performance of the proposed methodology and its generalisability in the plant identification domain, these are summarised in Table~\ref{tab:datasets}. Example images from each dataset are visible in Figure~\ref{fig:datasets}. Most datasets contain full-colour images, though ShapeCN consists of black and white images in which the boundary of the leaf is marked with a single pixel wide line and the 100-Leaves dataset contains black and white filled segmentations of leaf shapes.

\begin{table*}
    \centering
        \begin{tabular}{@{}lllR{1.3cm}R{1.7cm}R{1.5cm}R{1.3cm}@{}}
            \toprule
            Dataset                        & Image Type   & Condition     & Total Images  & Number of Species   & Total per Species  & Test per Species   \\ \midrule
            Flavia~\cite{flavia}           & Colour       & Isolated      & 1907          & 32                  & 50--77               & 10               \\
            Foliage~\cite{foliage}         & Colour       & Isolated      & 7200          & 60                  & 120                 & 20                \\
            Folio~\cite{folio}             & Colour       & Field/Isolated& 637           & 32                  & 18--20               & 4                \\
            Leafsnap Field~\cite{leafsnap} & Colour       & Field         & 7440          & 183                 & 10--145              & 5                \\
            Leafsnap Lab~\cite{leafsnap}   & Colour       & Lab (Pressed) & 22809         & 183                 & 40--268              & 10               \\
            ShapeCN~\cite{shapecn}         & Contour      & N/A           & 600           & 30                  & 20                  & 5                 \\
            Swedish Leaves~\cite{swedish}  & Colour       & Lab (Pressed) & 1125          & 15                  & 75                  & 15                \\
            100-Leaves~\cite{100leaves}    & Segmentation & N/A           & 1600          & 100                 & 16                  & 4                 \\
            \bottomrule
        \end{tabular}%
    \caption{Summary of publicly available leaf image datasets used in our experimental evaluation.\label{tab:datasets}}
\end{table*}

Conditions in which the images are taken vary significantly from one dataset to another. Some datasets contain colour images isolated against a flat white background, while others are taken in the field where leaves are placed on white paper in varying lighting conditions with shadows and blur often present, and some contain images of pressed leaves under lab conditions, typically on a back-lit surface. The size, number of species and number of images per species also varies widely between datasets, including large imbalances between classes. The number of test images per species has mostly been chosen to constitute approximately 20\% of the dataset as a whole, balanced equally between classes. In cases where some classes contain far fewer images than others, a test split less than 20\% is used to ensure all classes have a sufficient number of training examples.

Specific alterations have also been made for certain datasets. The original Leafsnap dataset includes 185 tree species, though field images are missing for \textit{Ulmus Procera}, and \textit{Quercus Falcata} contains insufficient lab images. These classes are therefore discarded entirely for the purposes of this study, leaving 183 tree species. \textit{Prunus Virgiana} and \textit{Prunus Sargentii} contained a number of incorrectly labelled images and \textit{Maclura Pomifera} contained duplicates of all lab images, these erroneous images were removed. Four field images of exceptionally low quality (such that no leaf boundary could be extracted) were also discarded.

The creators of the base ShapeCN dataset~\cite{shapecn} also provide two additional datasets, one of which contains 6 copies of each leaf image rotated by random angles between 0 and 360 degrees, totalling 3600 images. The other includes 4 copies of each leaf image scaled up in increments of 25\%, totalling 2400 images; hereafter these are referred to as ShapeCN-R and ShapeCN-S respectively.

\section{Methodology}\label{sec:impl}
This section describes the methodology we use for this investigation, including a detailed description of our proposed feature set. Firstly, the processes of segmentation and LAII extraction are described, including an explanation of the benefits of LAIIs in this context. Secondly, the proposed features are described, including basic shape features and those derived from the extracted LAIIs. Finally, the classification method used for evaluation is outlined.

\subsection{Leaf Segmentation}
The task of segmentation for the majority of datasets is trivial, some already contain segmentations or contours which can be used directly. Most others contain isolated leaves which can easily be segmented by a simple grey-level threshold. Lab and field images with variable backdrops and colour/size charts pose a more significant challenge.

Any charts present in the image are first removed in a preprocessing step. The maximum of the mean values of the pixels along each edge of the image is then used to determine the background saturation and grey-level which are then used as a basis for thresholding. A morphological closing operation is performed on the two resulting thresholds with a circular kernel of radius 5 pixels and the results summed to obtain a final segmentation for the image.

Once segmented, the stems are removed from the leaves as these drastically affect the LAII and vary randomly between leaves~\cite{beghin}. This is achieved by applying a morphological top-hat operation to the segmentation with a square kernel (sized proportionally to image resolution) and subtracting the result from the original segmentation. If the area of the segmentation would be reduced by more than 10\% by applying the top hat operation (e.g., for thin pine needles) then this operation is not applied.

Once the initial segmentation is obtained, all contours in the image are detected using the Suzuki-Abe algorithm~\cite{suzuki}. For isolated and pre-segmented images there is typically only a single contour which can be used directly. For lab and field images there may be a number of other contours present caused by irrelevant objects in the images or lighting effects.

To select the most likely candidate contour, the contour with the largest area, that has a length greater than a fixed minimum and that is within a reasonable distance (proportional to image resolution) from the centre of the image is used. This procedure eliminates most errors due to lighting effects (which typically occur at the edge of images) and any small irrelevant objects in the image. If no suitable contour is found then a fallback method using canny edge detection~\cite{canny} is used.

\subsection{LAII Extraction}\label{sec:LAIIex}
\begin{figure}[b]
    \centering
    \includegraphics[width=0.5\linewidth]{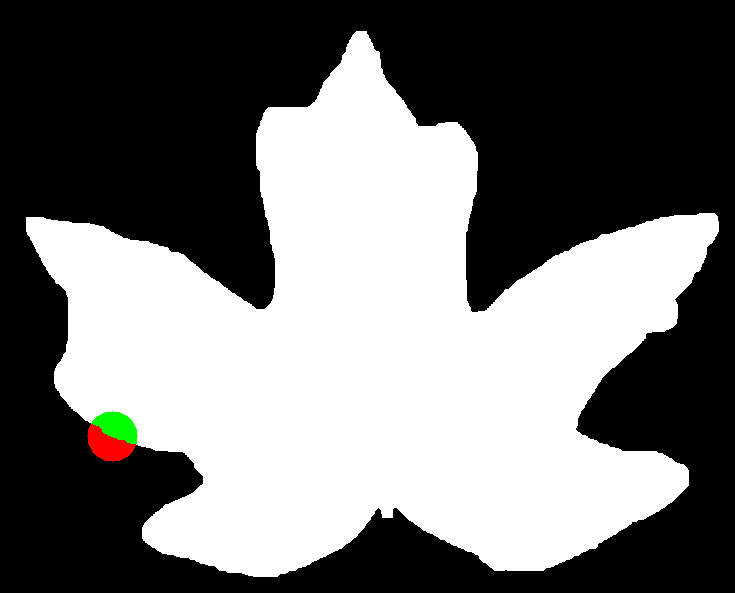}
    \caption{LAII extraction method; circular mask shown against segmentation image, with intersection in green and non-intersecting region in red. The mask is moved around the perimeter of the shape and the overlapping area, i.e., curvature recorded at regular intervals.\label{fig:cmap_ex}}
\end{figure}

LAIIs are extracted using the method presented in~\cite{integral}, which is also used in~\cite{leafsnap} and~\cite{costarica}. A small circular mask is moved around the edge of the shape and a bitwise-and operation applied to obtain the intersection of the mask with the segmentation. The curvature at a given point is approximated as the proportion of the circle which is filled (i.e., the ratio of the area of the intersection to the area of the entire mask). A visualisation of one step in this process is shown in Figure~\ref{fig:cmap_ex}.

A fixed number of points are selected around the shape boundary, 256 is chosen for this application based on cross-validation performance on the training set. LAIIs are extracted at multiple scales by varying the radius of the mask, the radius is defined as a percentage of the length of the contour to help retain scale invariance, this application makes use of 5 scales: 1\%, 2.5\%, 5\%, 10\% and 15\%, again selected based on cross-validation performance.

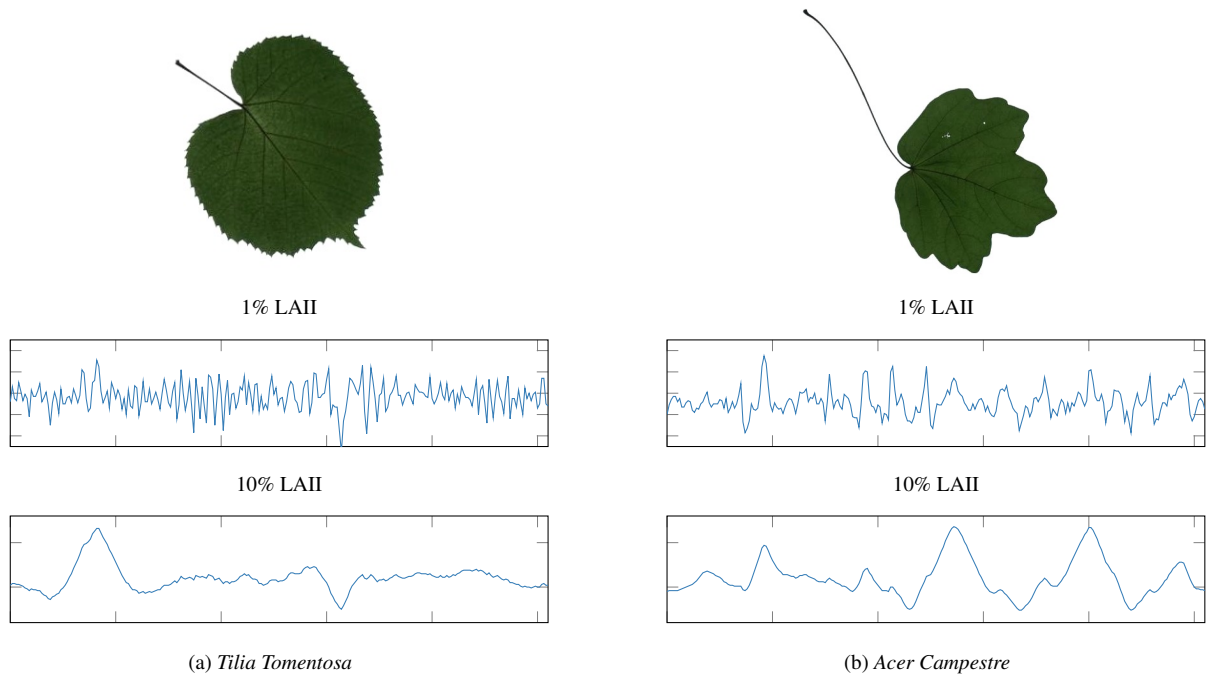
\begin{figure*}[!ht]
    \centering
    \subfloat[\textit{Tilia Tomentosa}\label{fig:curve_maps_a}]{%
        \parbox{0.5\textwidth}{%
            \centering
            \includegraphics[height=0.45\linewidth]{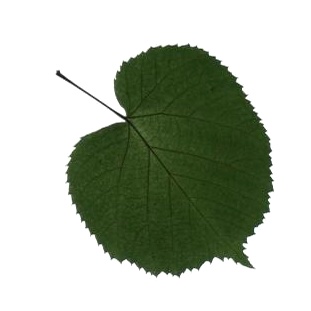}
            \\
            \begin{tikzpicture}
                \begin{axis}[
                    xmin=0,
                    xmax=255,
                    ymin=90,
                    ymax=190,
                    no markers,
                    height=3cm,
                    width=\linewidth,
                    xticklabels={,,},
                    yticklabels={,,},
                    title={\footnotesize{1\% LAII}}]
                    \addplot[color=Set1-B] coordinates {%
                        (0,132) (1,146) (2,135) (3,133) (4,150) (5,142) (6,133) (7,139) (8,135) (9,118) (10,143) (11,137) (12,137) (13,142) (14,149) (15,132) (16,136) (17,141) (18,129) (19,110) (20,131) (21,141) (22,134) (23,121) (24,144) (25,130) (26,130) (27,136) (28,137) (29,131) (30,144) (31,138) (32,131) (33,138) (34,162) (35,159) (36,135) (37,123) (38,126) (39,150) (40,152) (41,171) (42,165) (43,142) (44,131) (45,136) (46,142) (47,126) (48,150) (49,144) (50,135) (51,143) (52,138) (53,125) (54,135) (55,120) (56,143) (57,128) (58,126) (59,147) (60,145) (61,119) (62,131) (63,151) (64,132) (65,132) (66,144) (67,130) (68,135) (69,141) (70,134) (71,125) (72,140) (73,153) (74,143) (75,131) (76,127) (77,142) (78,150) (79,116) (80,129) (81,162) (82,140) (83,124) (84,134) (85,151) (86,128) (87,103) (88,154) (89,138) (90,116) (91,146) (92,128) (93,112) (94,157) (95,155) (96,144) (97,110) (98,135) (99,158) (100,139) (101,105) (102,136) (103,152) (104,120) (105,139) (106,140) (107,126) (108,132) (109,146) (110,143) (111,156) (112,129) (113,150) (114,139) (115,124) (116,148) (117,149) (118,123) (119,137) (120,125) (121,130) (122,125) (123,145) (124,156) (125,139) (126,145) (127,129) (128,135) (129,136) (130,123) (131,132) (132,145) (133,151) (134,148) (135,134) (136,131) (137,122) (138,123) (139,144) (140,151) (141,152) (142,121) (143,131) (144,159) (145,158) (146,136) (147,136) (148,139) (149,140) (150,153) (151,163) (152,120) (153,129) (154,127) (155,127) (156,106) (157,85) (158,114) (159,123) (160,137) (161,154) (162,152) (163,148) (164,143) (165,139) (166,149) (167,166) (168,123) (169,102) (170,133) (171,164) (172,151) (173,135) (174,113) (175,138) (176,140) (177,154) (178,123) (179,127) (180,138) (181,141) (182,146) (183,156) (184,138) (185,138) (186,129) (187,135) (188,140) (189,141) (190,135) (191,130) (192,148) (193,144) (194,141) (195,141) (196,140) (197,137) (198,136) (199,155) (200,135) (201,130) (202,134) (203,140) (204,150) (205,138) (206,134) (207,149) (208,133) (209,140) (210,143) (211,126) (212,139) (213,131) (214,146) (215,140) (216,136) (217,136) (218,134) (219,128) (220,140) (221,153) (222,120) (223,143) (224,147) (225,131) (226,112) (227,153) (228,134) (229,117) (230,137) (231,152) (232,137) (233,132) (234,121) (235,143) (236,156) (237,116) (238,135) (239,135) (240,127) (241,143) (242,131) (243,120) (244,145) (245,135) (246,140) (247,133) (248,139) (249,141) (250,126) (251,130) (252,154) (253,154) (254,130) (255,128)%
                    };
                \end{axis}
            \end{tikzpicture}
            \\
            \begin{tikzpicture}
                \begin{axis}[
                    xmin=0,
                    xmax=255,
                    ymin=60,
                    ymax=180,
                    no markers,
                    height=3cm,
                    width=\linewidth,
                    xticklabels={,,},
                    yticklabels={,,},
                    title={\footnotesize{10\% LAII}}]
                    \addplot[color=Set1-B] coordinates {%
                        (0,102) (1,104) (2,104) (3,103) (4,103) (5,102) (6,101) (7,99) (8,98) (9,96) (10,98) (11,97) (12,97) (13,96) (14,96) (15,93) (16,91) (17,90) (18,87) (19,86) (20,89) (21,90) (22,92) (23,93) (24,97) (25,99) (26,102) (27,106) (28,111) (29,115) (30,121) (31,126) (32,131) (33,137) (34,144) (35,148) (36,149) (37,151) (38,154) (39,160) (40,163) (41,166) (42,166) (43,161) (44,156) (45,152) (46,148) (47,142) (48,138) (49,133) (50,128) (51,123) (52,118) (53,113) (54,109) (55,105) (56,104) (57,100) (58,97) (59,97) (60,96) (61,93) (62,94) (63,95) (64,93) (65,94) (66,95) (67,94) (68,95) (69,96) (70,97) (71,97) (72,100) (73,102) (74,102) (75,102) (76,103) (77,105) (78,107) (79,105) (80,108) (81,112) (82,110) (83,109) (84,110) (85,111) (86,110) (87,108) (88,113) (89,112) (90,111) (91,112) (92,111) (93,110) (94,114) (95,113) (96,112) (97,108) (98,109) (99,110) (100,107) (101,104) (102,106) (103,106) (104,104) (105,106) (106,106) (107,107) (108,109) (109,112) (110,113) (111,115) (112,113) (113,114) (114,111) (115,109) (116,109) (117,108) (118,104) (119,104) (120,102) (121,103) (122,104) (123,107) (124,108) (125,108) (126,108) (127,106) (128,107) (129,107) (130,108) (131,110) (132,114) (133,116) (134,116) (135,115) (136,115) (137,115) (138,116) (139,120) (140,122) (141,123) (142,121) (143,122) (144,123) (145,122) (146,118) (147,115) (148,111) (149,108) (150,105) (151,100) (152,93) (153,89) (154,84) (155,81) (156,77) (157,75) (158,79) (159,84) (160,89) (161,96) (162,101) (163,104) (164,107) (165,110) (166,112) (167,114) (168,110) (169,109) (170,111) (171,114) (172,112) (173,109) (174,106) (175,108) (176,108) (177,109) (178,107) (179,108) (180,110) (181,111) (182,113) (183,114) (184,112) (185,111) (186,110) (187,110) (188,111) (189,112) (190,112) (191,113) (192,115) (193,116) (194,116) (195,115) (196,115) (197,115) (198,114) (199,115) (200,113) (201,112) (202,111) (203,111) (204,113) (205,111) (206,111) (207,113) (208,111) (209,112) (210,114) (211,113) (212,115) (213,116) (214,118) (215,119) (216,119) (217,119) (218,119) (219,118) (220,119) (221,120) (222,118) (223,119) (224,118) (225,116) (226,114) (227,116) (228,113) (229,111) (230,112) (231,113) (232,111) (233,109) (234,108) (235,109) (236,108) (237,105) (238,105) (239,104) (240,103) (241,103) (242,101) (243,100) (244,101) (245,100) (246,100) (247,99) (248,100) (249,101) (250,100) (251,100) (252,103) (253,104) (254,102) (255,102)%
                    };
                \end{axis}
            \end{tikzpicture}%
    }}
    \hfill
    \subfloat[\textit{Acer Campestre}\label{fig:curve_maps_b}]{%
        \parbox{0.5\textwidth}{%
            \centering
            \includegraphics[height=0.45\linewidth]{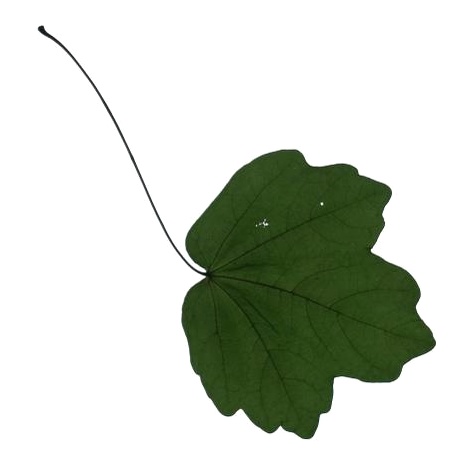}
            \\
            \begin{tikzpicture}
                \begin{axis}[
                    xmin=0,
                    xmax=255,
                    ymin=90,
                    ymax=190,
                    no markers,
                    height=3cm,
                    width=\linewidth,
                    xticklabels={,,},
                    yticklabels={,,},
                    title={\footnotesize{1\% LAII}}]
                    \addplot[color=Set1-B] coordinates {%
                        (0,120) (1,129) (2,134) (3,137) (4,137) (5,132) (6,135) (7,127) (8,127) (9,131) (10,133) (11,132) (12,124) (13,122) (14,126) (15,125) (16,130) (17,132) (18,137) (19,141) (20,131) (21,127) (22,130) (23,130) (24,130) (25,134) (26,133) (27,126) (28,135) (29,122) (30,126) (31,132) (32,126) (33,130) (34,138) (35,149) (36,113) (37,103) (38,107) (39,114) (40,130) (41,125) (42,124) (43,126) (44,139) (45,165) (46,175) (47,168) (48,147) (49,129) (50,123) (51,131) (52,127) (53,125) (54,126) (55,121) (56,127) (57,132) (58,135) (59,134) (60,129) (61,121) (62,130) (63,141) (64,140) (65,141) (66,144) (67,132) (68,136) (69,135) (70,140) (71,130) (72,130) (73,135) (74,131) (75,114) (76,132) (77,158) (78,139) (79,126) (80,132) (81,134) (82,128) (83,125) (84,139) (85,137) (86,135) (87,129) (88,118) (89,117) (90,117) (91,108) (92,129) (93,158) (94,161) (95,160) (96,135) (97,128) (98,127) (99,122) (100,129) (101,127) (102,112) (103,121) (104,120) (105,115) (106,160) (107,165) (108,143) (109,129) (110,133) (111,139) (112,133) (113,120) (114,114) (115,114) (116,112) (117,117) (118,127) (119,122) (120,122) (121,134) (122,146) (123,165) (124,137) (125,110) (126,107) (127,120) (128,130) (129,129) (130,129) (131,133) (132,137) (133,143) (134,139) (135,150) (136,154) (137,151) (138,142) (139,139) (140,137) (141,131) (142,128) (143,124) (144,124) (145,130) (146,125) (147,130) (148,134) (149,135) (150,131) (151,123) (152,128) (153,122) (154,119) (155,118) (156,123) (157,136) (158,143) (159,144) (160,142) (161,147) (162,137) (163,132) (164,126) (165,123) (166,121) (167,105) (168,112) (169,119) (170,128) (171,126) (172,118) (173,123) (174,135) (175,136) (176,132) (177,136) (178,141) (179,155) (180,145) (181,134) (182,117) (183,125) (184,122) (185,119) (186,128) (187,134) (188,131) (189,121) (190,131) (191,123) (192,131) (193,128) (194,134) (195,141) (196,139) (197,133) (198,128) (199,133) (200,161) (201,162) (202,151) (203,137) (204,133) (205,133) (206,139) (207,131) (208,112) (209,123) (210,124) (211,128) (212,143) (213,142) (214,142) (215,139) (216,129) (217,113) (218,123) (219,114) (220,103) (221,124) (222,132) (223,140) (224,131) (225,128) (226,131) (227,133) (228,144) (229,156) (230,130) (231,111) (232,116) (233,115) (234,124) (235,131) (236,131) (237,128) (238,139) (239,134) (240,138) (241,138) (242,144) (243,147) (244,145) (245,153) (246,147) (247,139) (248,120) (249,114) (250,118) (251,126) (252,133) (253,135) (254,132) (255,125)%
                    };
                \end{axis}
            \end{tikzpicture}
            \\
            \begin{tikzpicture}
                \begin{axis}[
                    xmin=0,
                    xmax=255,
                    ymin=60,
                    ymax=180,
                    no markers,
                    height=3cm,
                    width=\linewidth,
                    xticklabels={,,},
                    yticklabels={,,},
                    title={\footnotesize{10\% LAII}}]
                    \addplot[color=Set1-B] coordinates {%
                        (0,95) (1,96) (2,96) (3,96) (4,96) (5,96) (6,97) (7,98) (8,99) (9,100) (10,101) (11,102) (12,103) (13,105) (14,107) (15,110) (16,113) (17,116) (18,117) (19,118) (20,117) (21,116) (22,115) (23,113) (24,112) (25,111) (26,109) (27,106) (28,105) (29,103) (30,102) (31,102) (32,101) (33,101) (34,101) (35,101) (36,97) (37,96) (38,98) (39,102) (40,108) (41,115) (42,122) (43,129) (44,136) (45,144) (46,147) (47,146) (48,141) (49,135) (50,130) (51,126) (52,122) (53,119) (54,117) (55,114) (56,114) (57,114) (58,113) (59,112) (60,112) (61,110) (62,111) (63,113) (64,113) (65,114) (66,114) (67,112) (68,111) (69,110) (70,109) (71,107) (72,106) (73,106) (74,106) (75,105) (76,108) (77,110) (78,108) (79,107) (80,105) (81,104) (82,103) (83,101) (84,101) (85,100) (86,98) (87,96) (88,95) (89,97) (90,100) (91,104) (92,110) (93,117) (94,120) (95,121) (96,116) (97,113) (98,108) (99,105) (100,102) (101,99) (102,96) (103,96) (104,96) (105,95) (106,100) (107,100) (108,97) (109,93) (110,91) (111,88) (112,83) (113,79) (114,76) (115,75) (116,76) (117,79) (118,84) (119,90) (120,96) (121,101) (122,106) (123,112) (124,113) (125,114) (126,118) (127,123) (128,128) (129,134) (130,140) (131,145) (132,152) (133,157) (134,162) (135,166) (136,168) (137,167) (138,165) (139,161) (140,157) (141,151) (142,147) (143,141) (144,136) (145,131) (146,126) (147,121) (148,115) (149,111) (150,106) (151,101) (152,98) (153,95) (154,94) (155,93) (156,94) (157,96) (158,97) (159,96) (160,93) (161,91) (162,88) (163,85) (164,82) (165,79) (166,76) (167,74) (168,74) (169,76) (170,79) (171,81) (172,84) (173,88) (174,93) (175,98) (176,102) (177,105) (178,107) (179,110) (180,108) (181,105) (182,102) (183,101) (184,101) (185,101) (186,104) (187,107) (188,111) (189,114) (190,119) (191,125) (192,130) (193,135) (194,140) (195,145) (196,150) (197,154) (198,159) (199,162) (200,167) (201,167) (202,164) (203,159) (204,154) (205,147) (206,142) (207,135) (208,128) (209,125) (210,121) (211,117) (212,115) (213,110) (214,104) (215,97) (216,91) (217,84) (218,80) (219,76) (220,74) (221,75) (222,77) (223,81) (224,83) (225,85) (226,89) (227,92) (228,95) (229,97) (230,95) (231,94) (232,95) (233,97) (234,100) (235,103) (236,107) (237,109) (238,114) (239,117) (240,121) (241,123) (242,126) (243,128) (244,128) (245,127) (246,121) (247,115) (248,109) (249,104) (250,100) (251,98) (252,98) (253,97) (254,97) (255,96) %
                    };
                \end{axis}
            \end{tikzpicture}%
    }}
    \caption{Leaf image and extracted LAIIs at small (1\%) and large (10\%) scales for two leaves from the Leafsnap Lab dataset. The difference in leaf margin characteristics are clearly visible in the 1\% LAII, while the overall leaf shape is captured effectively in the 10\% LAII.\label{fig:curve_maps}}
\end{figure*}

Examples of LAIIs for two leaf types can be seen in Figure~\ref{fig:curve_maps}, the first at the 1\% scale and the second at 10\%. Figure~\ref{fig:curve_maps_a} shows a leaf with a serrated margin, but a relatively simple overall shape, while Figure~\ref{fig:curve_maps_b} shows a leaf with a smooth margin, but more complex overall shape.

These differences are reflected in the LAIIs at both scales, with the 1\% LAII for Figure~\ref{fig:curve_maps_a} containing lots of high-frequency activity, while its counterpart in~\ref{fig:curve_maps_b} contains noticeably less. Meanwhile, the 10\% LAII for~\ref{fig:curve_maps_a} contains little low-frequency activity and some low amplitude, high-frequency activity, while for~\ref{fig:curve_maps_b} contains noticeably more low-frequency activity and almost no high-frequency activity. Looking closely, the exact shape of each leaf can be seen to correspond directly to each LAII\@.

At low resolutions, the LAIIs begin to be affected by pixel level artefacts and are prone to large amounts of noise. For this reason, the 1\% and 2.5\% scales are discarded for the ShapeCN dataset which contains $256\times256$ pixel images, improving accuracy by over 5\%. This is not true for higher resolution images for which these smaller scales aid performance significantly.

\subsection{Feature Extraction}
Once a valid contour has been obtained, we propose a number of features to be extracted which concisely represent the discriminating aspects of the leaf shape. These fall into two broad categories: features based on the overall shape of the leaf and features based on the extracted LAIIs\@. LAIIs are 1D signals similar to audio or ECG/EEG signals. Consequently, many of the LAII features proposed are inspired by feature extraction techniques typically used for these more common 1D signals.

Our LAII-based features differ from the histograms of curvature scale (HoCS) introduced in~\cite{leafsnap} as, where HoCS derive histograms from the magnitude of LAIIs directly, we instead focus on the frequency spectra of LAIIs. As discussed in relation to Figure~\ref{fig:curve_maps} above, we propose that the frequency spectrum of a LAII is a better discriminator than a histogram of magnitudes.

\subsubsection{Basic Shape Features}~\\
Four basic shape features are extracted, selected due to their prevalence in existing work on leaf identification. \emph{Solidity}, the ratio of the contour area to the area of the convex hull of the contour. \emph{Circularity}, the ratio of the area of the contour to the area of a circle with equal perimeter to the length of the contour. \emph{Rectangularity}, the ratio of the area of the contour to the area of the minimum-area rectangle containing the contour and \emph{Compactness}, the ratio of the length of the contour to the are of the contour.


\subsubsection{LAII Features}~\\
The rest of the proposed features are extracted from the LAIIs at the 5 different scales, as described in Section~\ref{sec:LAIIex}. These features are predominantly statistical, as well focusing on the frequency spectra of the LAIIs, as is common for audio and ECG/EEG data. These features are described below.

\paragraph{Basic Statistical}~\\
The mean and standard deviation of each LAII, first and second differences, and absolute first and second differences are used. The area under the curve is also included to give some indication of the peaks present in the LAII\@.

\paragraph{Bending Energy}~\\
The bending energy, $B$, denotes the energy stored in a shape and has been used for leaf identification before~\cite{leafprocessor}. It is defined as the mean value of the squared entries in a sequence, $k_i$, of length $K$.
\[
    B = \frac{1}{K}\sum_{i=1}^{K} k_i^2
\]
\paragraph{Signal Entropy}~\\
Signal entropy is also used as a feature, calculated from a 128-bin histogram of each LAII\@.
\paragraph{Frequency (FFT)}~\\
The 256-bin real-valued FFT of the LAII, from this the spectral centroid, a feature typically used for audio, is calculated as shown. $f_i$ being the frequency for a given FFT bin, and $k_i$ the corresponding value for that bin.
\[
    \mathit{centroid} = \frac{\sum_{i=1}^{K} f_{i}k_{i}}{\sum_{i=1}^{K} k_{i}}
\]
The centroid and full, normalised FFT for each LAII are also included in the feature set. FFT features are used due to their effective encoding of the frequency spectra of the LAII\@.

\subsection{Evaluation}
The full feature set contains 719 individual features; PCA is used to reduce the dimensionality to 128. This vastly improves training times with almost no effect on classification performance.

Classification is performed using an SVM with an RBF kernel ($C=1000$, $\gamma=7$) and balanced class weights. ScikitLearn's~\cite{scikit-learn} SVM one-vs-one multi-class classification is used. Hyper-parameters are selected using 5-fold cross-validation on the training sets to provide good overall performance on all datasets; marginally improved results could be achieved by tuning hyper-parameters for each dataset individually.

\section{Experimental Results}\label{sec:res}
This section provides a comprehensive evaluation of our proposed feature set for the task of leaf identification. We include classification results for all the standard leaf datasets and a comparison to the current state-of-the-art. The invariance of our method to scale and rotation invariance is also demonstrated, along with an assessment of the generalisability of the feature set for generic shape matching tasks. Finally, the computational efficiency of the feature extraction and classification processes are explored.

\subsection{Leaf Identification}
The proposed methodology is first evaluated on the standard leaf datasets described in Table~\ref{tab:datasets}. The classification performance---in terms of precision, recall and F1 score---for each dataset is given in Table~\ref{tab:all_res}.

\begin{table*}
    \centering
    \resizebox{\linewidth}{!}{%
        \begin{tabular}{@{}lR{1.7cm}R{1.7cm}R{1.7cm}R{1.7cm}R{1.7cm}R{1.7cm}R{1.7cm}R{1.7cm}@{}}
            \toprule
            & Flavia & Foliage & Folio & Leafsnap Field & Leafsnap Lab & ShapeCN  & Swedish Leaves & 100-Leaves \\ \midrule
            \multicolumn{1}{@{}l}{Recall}       & 0.966  & 0.931   & 0.922 & 0.649          & 0.924        & 0.967    & 0.978          & 0.910      \\
            \multicolumn{1}{@{}l}{Precision}    & 0.970  & 0.933   & 0.938 & 0.696          & 0.928        & 0.972    & 0.979          & 0.924      \\
            \multicolumn{1}{@{}l}{F1-Score}     & 0.965  & 0.929   & 0.918 & 0.636          & 0.922        & 0.965    & 0.978          & 0.908      \\
            \bottomrule
        \end{tabular}%
    }
    \caption{Classification performance for standard leaf datasets.\label{tab:all_res}}
\end{table*}

All datasets apart from Leafsnap Field achieve over 90\% recall and precision for the first prediction, with the highest being Swedish Leaves at 97.8\% recall. Leafsnap Field has a worse performance, with just 64.9\% of first predictions correct. This is likely due to the poor quality of the input images, with large disruption caused by blur and shadows resulting in inaccurate segmentations.

The higher quality field images of Folio have caused much less of an issue, demonstrating that our method is robust to some level of shadow, but deals poorly with extreme lighting conditions and blur. An improved segmentation technique may aid performance in this case, though it is certainly clear that a high quality segmentation is required for effective LAII extraction.

Table~\ref{tab:sota} shows detailed comparisons between the classification accuracies of our method and state-of-the-art methods for all datasets. The Leafsnap dataset is excluded due to the incomparability of previous evaluations; existing methods use the field and lab datasets combined. As shown, our method's performance is consistently comparable to existing approaches that utilise complex feature sets and, when compared with shape-only state-of-the-art approaches, our method performs significantly better. Many state-of-the-art methods make no demonstration of scale or rotation invariance meaning that results may not be representative of deployed performance. This is particularly true for shape and texture features which are often heavily orientation and scale dependent.

\begin{table}
    \centering
    \resizebox{\linewidth}{!}{%
    \begin{tabular}{@{}lrrr@{}}
        \toprule
        Dataset & Our Performance & SotA Method &  SotA Performance   \\ \midrule
        Flavia         & 0.966 & LeafNet~\cite{leafnet} 				  & 0.979   \\
                       &       & Shape, Texture \& Colour~\cite{glcmshen} & 0.972 	\\
                       &	   & Shape-only~\cite{glcmshen} 			  & 0.768   \\
        Foliage        & 0.931 & LeafNet~\cite{leafnet} 				  & 0.958   \\
                       &       & Shape, Texture \& Colour~\cite{glcmshen} & 0.950   \\
                       &	   & Shape-only~\cite{glcmshen} 			  & 0.698   \\
        Folio          & 0.914 & GoogleNet~\cite{bagofwords} 			  & 0.976   \\
                       &       & HoG-based~\cite{bagofwords} 			  & 0.928   \\
        ShapeCN		   & 0.953 & Shape-only~\cite{shapecn}   			  & 0.838   \\
        Swedish Leaves & 0.978 & Shape \& Texture~\cite{centrist}         & 0.979   \\
                       &       & Shape-only~\cite{centrist}    			  & 0.908   \\
        100-Leaves     & 0.910 & Shape \& Texture~\cite{100leaves}        & 0.968   \\
                       &       & Shape-only~\cite{kadir2015leaf} 		  & 0.880   \\
        \bottomrule
    \end{tabular}%
    }
    \caption{Comparison between our proposed method and state-of-the-art (SotA) methods for each dataset, in terms of classification accuracy. Our method often performs comparably to SotA methods using more complex feature sets and outperforms the SotA shape-only methods for all datasets.\label{tab:sota}}
\end{table}

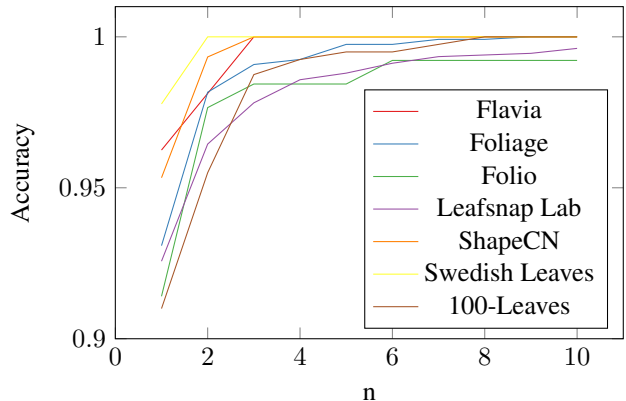
\begin{figure}
    \centering
    \begin{tikzpicture}
        \begin{axis}[
            height=6cm,
            width=\linewidth,
            xlabel=n,
            ylabel=Accuracy,
            xmin=0,
            xmax=11,
            ymin=0.9,
            legend pos=south east,
            no markers,
            cycle multi list={Set1-7}]
            \addplot coordinates {%
                (1, 0.9625) (2, 0.98125) (3, 1.0) (4, 1.0) (5, 1.0) (6, 1.0) (7, 1.0) (8, 1.0) (9, 1.0) (10, 1.0)
            };
            \addlegendentry{Flavia}
            \addplot coordinates {%
                (1, 0.9308333333333333) (2, 0.9816666666666667) (3, 0.9908333333333333) (4, 0.9925) (5, 0.9975) (6, 0.9975) (7, 0.9991666666666666) (8, 0.9991666666666666) (9, 1.0) (10, 1.0)
            };
            \addlegendentry{Foliage}
            \addplot coordinates {%
                (1, 0.9140625) (2, 0.9765625) (3, 0.984375) (4, 0.984375) (5, 0.984375) (6, 0.9921875) (7, 0.9921875) (8, 0.9921875) (9, 0.9921875) (10, 0.9921875)
            };
            \addlegendentry{Folio}
            \addplot coordinates {%
                (1, 0.9256830601092896) (2, 0.9644808743169399) (3, 0.9781420765027322) (4, 0.985792349726776) (5, 0.9879781420765027) (6, 0.9912568306010929) (7, 0.9934426229508196) (8, 0.9939890710382514) (9, 0.994535519125683) (10, 0.9961748633879781)
            };
            \addlegendentry{Leafsnap Lab}
            \addplot coordinates {%
                (1, 0.9533333333333334) (2, 0.9933333333333333) (3, 1.0) (4, 1.0) (5, 1.0) (6, 1.0) (7, 1.0) (8, 1.0) (9, 1.0) (10, 1.0)
            };
            \addlegendentry{ShapeCN}
            \addplot coordinates {%
                (1, 0.9777777777777777) (2, 1.0) (3, 1.0) (4, 1.0) (5, 1.0) (6, 1.0) (7, 1.0) (8, 1.0) (9, 1.0) (10, 1.0)
            };
            \addlegendentry{Swedish Leaves}
            \addplot coordinates {%
                (1, 0.91) (2, 0.955) (3, 0.9875) (4, 0.9925) (5, 0.995) (6, 0.995) (7, 0.9975) (8, 1.0) (9, 1.0) (10, 1.0)
            };
            \addlegendentry{100-Leaves}
        \end{axis}
    \end{tikzpicture}
    \caption{Top-n accuracy scores for standard leaf datasets (excluding Leafsnap Field).\label{fig:all_res}}
\end{figure}

Figure~\ref{fig:all_res} demonstrates how, while first prediction accuracy is often in the low nineties, it rapidly improves (4\% on average between top-1 and top-2 accuracy) and all datasets achieve over 98\% accuracy for the top-4 predictions. So, while a few very similar leaf species might be confused, the presented methodology effectively separates these closely related species from the many other species present in the dataset. Only two of the seven included datasets fail to achieve 100\% accuracy after 9 predictions.

The effectiveness of the proposed LAII-based features is demonstrated in Table~\ref{tab:feature_eff}. In all cases the LAII-based features significantly outperform the basic shape features in isolation. When these basic shape features are used in conjunction with the LAII-based features a small benefit to accuracy is observed, though in the case of the Folio dataset the opposite is seen. This is likely due to the similarity in overall shape of leaves in Folio resulting in the basic shape features in fact causing confusion between classes rather than discriminating between them.

\begin{table}
    \centering
        \resizebox{\linewidth}{!}{%
    \begin{tabular}{@{}lrrr@{}}
        \toprule
        Dataset & Basic Shape & LAII-based & Full Feature Set  \\ \midrule
        \multicolumn{1}{@{}l}{Flavia}         & 0.756  & 0.950 & 0.966  \\
        \multicolumn{1}{@{}l}{Foliage}        & 0.618  & 0.927 & 0.931  \\
        \multicolumn{1}{@{}l}{Folio}          & 0.609  & 0.930 & 0.914  \\
        \multicolumn{1}{@{}l}{Leafsnap Field} & 0.302  & 0.635 & 0.649  \\
        \multicolumn{1}{@{}l}{Leafsnap Lab}   & 0.492  & 0.914 & 0.924  \\
        \multicolumn{1}{@{}l}{ShapeCN}        & 0.807  & 0.940 & 0.953  \\
        \multicolumn{1}{@{}l}{Swedish Leaves} & 0.831  & 0.973 & 0.978  \\
        \multicolumn{1}{@{}l}{100-Leaves}     & 0.613  & 0.908 & 0.910  \\
        \bottomrule
    \end{tabular}%
    }
    \caption{Classification accuracies for all dataset when using only basic shape features, only LAII-based features, and the complete feature set.\label{tab:feature_eff}}
\end{table}

No datasets including images of leaves of different colours for each species are currently available. As the proposed method utilises only shape features, any form of dataset generation by altering the hue of images from the currently available datasets would not result in any changes to the performance. In any case, segmentation would be the only affected aspect, which is not the primary focus of this paper.

\subsection{Scale and Rotation Invariance}
In order to evaluate the invariance to scale and rotation of the proposed methodology, an SVM is trained on the entire base ShapeCN dataset (600 images), this is then used to make predictions for the entire ShapeCN-R (3600 images) and ShapeCN-S (2400 images). As the ShapeCN-R and ShapeCN-S datasets only contain processed versions of the images contained in ShapeCN, this imbalance of training to testing samples is reasonable.

The results of this evaluation are shown in Table~\ref{tab:rot_scale}, along with those for the standard dataset, as above, for comparison. The model performs comparably or better for both the rotated (98\% recall) and scaled (97\% recall) datasets indicating that this method is indeed invariant to scale and rotation of shapes within the images being classified. This is as expected based on the inherent scale and rotation invariance of the extracted LAII\@.

The images in the ShapeCN-S dataset are provided at $512\times512$ pixels, while ShapeCN and ShapeCN-R both contain $256\times256$ pixel images. Using the larger ShapeCN-S images directly results in significantly reduced performance (~10--15\%), rescaling the images to $256\times256$ pixels eliminates this performance loss, demonstrating that while our method is largely invariant to the scale of the leaf within the image, the resolution of the input image does impact performance.

\begin{table}
    \centering
    \begin{tabular}{@{}lrrr@{}}
        \toprule
        & Standard   & Rotated   & Scaled    \\ \midrule
        \multicolumn{1}{@{}l}{Recall}      & 0.967      & 0.980     & 0.967     \\
        \multicolumn{1}{@{}l}{Precision}   & 0.972      & 0.980     & 0.970     \\
        \multicolumn{1}{@{}l}{F1-Score}    & 0.965      & 0.980     & 0.968     \\
        \bottomrule
    \end{tabular}
    \caption{Classification performance when trained on Standard ShapeCN dataset and tested on Rotated ShapeCN-R and Scaled ShapeCN-S datasets.\label{tab:rot_scale}}
\end{table}

\subsection{Generalisability}
\begin{figure}
    \centering
    \subfloat[ShapeCN Fish]{%
        \includegraphics[height=0.14\linewidth]{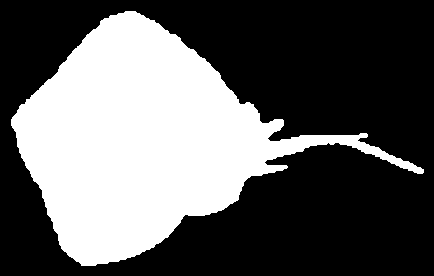}
        \includegraphics[height=0.14\linewidth]{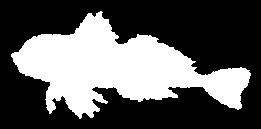}
    }
    \hfill
    \subfloat[MPEG-7 CE]{%
        \includegraphics[height=0.14\linewidth]{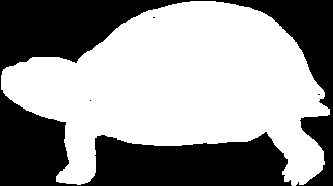}
        \includegraphics[height=0.14\linewidth]{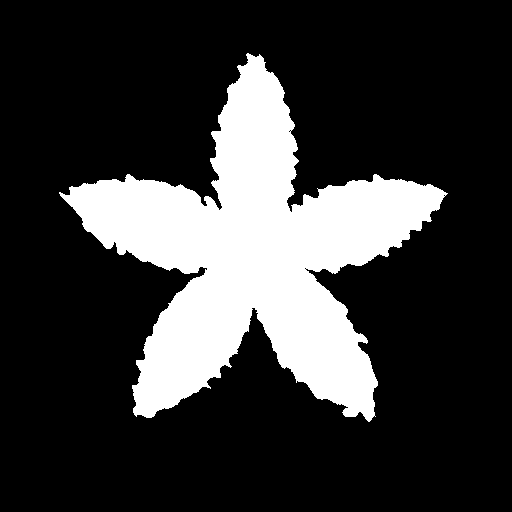}
    }
    \caption{Example images from generic shape image datasets.\label{fig:gen_datasets}}
\end{figure}

To evaluate the generalisability of the proposed features, we test the performance of our methodology on two other generic shape datasets. Example images from each dataset are shown in Figure~\ref{fig:gen_datasets}.

In addition to leaf images, Backes \emph{et al.}~\cite{shapecn} provide a set of 11000 black and white segmentation images of fish. 1100 classes of fish are included with 10 images per fish, these 10 are scaled (in increments of 25\%) and randomly rotated versions of a single image. As such, a test set of 5 images per class is selected, leaving a training set containing 5 images per class.

The MPEG-7 Core Experiment dataset~\cite{mpeg7} is another public dataset for shape classification which includes 1400 black and white segmentation images split between 70 classes, each class contains 20 images. The images in the MPEG-7 dataset are significantly more diverse than the ShapeCN fish dataset; for this evaluation, a test set containing 5 images per class is randomly selected.

To classify images from these generic shape datasets the same extraction and classification procedure is used as for the leaf datasets, with results given in Table~\ref{tab:fish}. The high recall rates (99\% and 98\%) on these generic shape datasets indicate that the proposed LAII-based methodology provides a strong candidate for generic shape recognition tasks, providing accurate segmentation is possible. 

\begin{table}
    \centering
    \begin{tabular}{@{}lrr@{}}
        \toprule
        & ShapeCN Fish   & MPEG-7   \\ \midrule
        \multicolumn{1}{@{}l}{Recall}      & 0.989          & 0.974    \\
        \multicolumn{1}{@{}l}{Precision}   & 0.991          & 0.981    \\
        \multicolumn{1}{@{}l}{F1-Score}    & 0.989          & 0.973    \\
        \bottomrule
    \end{tabular}
    \caption{Generic shape classification performance.\label{tab:fish}}
\end{table}

\subsection{Computational Efficiency}\label{sec:eff}
As the LAII extraction process is pixel driven, extraction time is resolution-dependant and can vary significantly between datasets. For lower resolution images, such as those of the Foliage dataset~\cite{foliage}, LAII extraction takes approximately 116ms per image. While for larger images, such as those in the Flavia dataset~\cite{flavia}, this process can take up to 1s per image. Once the LAIIs have been extracted, feature extraction time is minimal at approximately 14ms per image.\footnote{Timing measurements are taken on a 2.5 GHz Intel Core i7 2017 Apple MacBook Pro, averaged across 480 images.}

While extraction can be relatively slow, the extracted features can be stored with minimal space requirements compared with the original images (6\textsc{kb} per item). Even for larger datasets containing high resolution images, this process is still much quicker than training a large CNN such as LeafNet~\cite{leafnet}; the proposed method taking just a few hours in total, compared with as much as 32 hours to train the LeafNet CNN\@. The proposed method also proves performant on smaller datasets such as Folio and ShapeCN, where a large CNN would struggle to learn effectively without any pre-training.

There is significant room for optimisation of the LAII extraction process, such as that suggested in~\cite{leafsnap} by exploiting overlap in the masked area between frames. This could significantly improve the performance of the LAII extraction process, the clear bottleneck in the methodology proposed here.

Training of the SVM model takes from less than a second for smaller datasets, such as ShapeCN, and up to 2 minutes for the largest dataset, Leafsnap Lab, negligible compared with extraction time. The resulting models have significantly lower storage requirements than, for example, trained LeafNet models. The compressed SVM models require as little as 300\textsc{kb} for smaller datasets, with a maximum of 22\textsc{mb} for Leafsnap Lab. Again this is far smaller than a trained LeafNet model at 150\textsc{mb} (independent of dataset size).

These factors show that this method provides improvements both in terms of updatability, due to reduced training times, as well as scope for mobile deployment, as a result of significantly reduced model sizes still with reasonable feature extraction and classification times ($\ll1$ second).

\section{Conclusions \& Future Work}
This paper has presented a novel feature set for shape only leaf identification based on basic shape and LAII signal features. This methodology was evaluated on a number of publicly available leaf datasets with comparable results to existing methods which use colour and texture features in addition to shape. Over 90\% accuracy was achieved on all datasets excluding Leafsnap Field, with top-four accuracy over 98\%. Rotation and scale invariance of the feature extraction process was demonstrated, along with evaluation of the generalisability of the approach; achieving 91\% accuracy on the MPEG-7 Core Experiment dataset of generic shapes.

While not a particular focus for this paper, segmentation is likely the reason for poorer performance on field datasets, where blur and shadows cause particular problems. Further effort on improving segmentation of these images would likely bring field dataset performance closer to that shown for lab datasets. Stem removal as presented here is also not optimal, with particular problems cause for compound leaves---those consisting of multiple leaflets connected by a thin stem. Some work has aimed to tackle this problem~\cite{leafsnap,costarica}, but this typically reduces segmentation efficiency with little to no impact on performance. Finally, as mentioned in Section~\ref{sec:eff}, our LAII extraction implementation could be further optimised, this would have a major impact in reducing feature extraction time.

\section*{Acknowledgements}
Excluded for anonymous review.

{\small
\bibliographystyle{ieee}
\bibliography{bibliography}
}

\end{document}